%% file: egbib.tex
\documentclass[10pt,twocolumn,letterpaper]{article}

\usepackage{times}
\usepackage{epsfig}
\usepackage{graphicx}
\usepackage{amsmath}
\usepackage{amssymb}
\usepackage{bbm}
\usepackage{booktabs}
\usepackage{lipsum}
\usepackage{pifont}

\graphicspath{ {images/} }

\usepackage[pagebackref=true,breaklinks=true,letterpaper=true,colorlinks,bookmarks=false]{hyperref}

\usepackage{cvpr}

\cvprfinalcopy 

\def\cvprPaperID{12} 

\newcommand{\cmark}{\ding{51}}%
\newcommand{\xmark}{\ding{55}}%

\def\cvprPaperID{}
\begin{document}

\title{LCDet: Low-Complexity Fully-Convolutional Neural Networks for\\ Object Detection in Embedded Systems}

\author{Subarna Tripathi\\
UC San Diego
\footnotemark 
\\
{\tt\small stripathi@ucsd.edu}
\and
Gokce Dane\\
Qualcomm Inc.\\
{\tt\small gokced@qti.qualcomm.com}
\and
Byeongkeun Kang\\
UC San Diego\\
{\tt\small bkkang@ucsd.edu}
\and
Vasudev Bhaskaran\\
Qualcomm Inc.\\
{\tt\small vasudevb@qti.qualcomm.com}
\and
Truong Nguyen\\
UC San Diego\\
{\tt\small tqn001@eng.ucsd.edu}
}

\maketitle

\renewcommand*{\thefootnote}{\fnsymbol{footnote}}
\footnotetext[1]{Work done in part during an internship at Qualcomm.}
\renewcommand*{\thefootnote}{\arabic{footnote}}
\setcounter{footnote}{0}

\input{abstract.tex}

\input{introduction.tex}

\input{related_work.tex}

\input{methods.tex}

\input{results.tex}

\input{conclusions.tex}
\input{ack.tex}

{\small
\sloppy
\bibliographystyle{ieee}
\bibliography{egbib}
}

\end{document}

%% file: abstract.tex
\begin{abstract} 
\label{sec:abstract}

Deep Convolutional Neural Networks (CNN) are the state-of-the-art performers for the object detection task. It is well known that object detection requires more computation and memory than image classification. 
In this work, we propose \emph{LCDet}, a fully-convolutional neural network for generic object detection that aims to work in embedded systems.
We design and develop an end-to-end TensorFlow(TF)-based 
model.
The detection works by a single forward pass through the network.
Additionally, we employ 8-bit quantization on the learned weights. 
As a use case, we choose face detection and train the proposed model on images containing a varying number of faces of different sizes.  
We evaluate the face detection performance on publicly available dataset FDDB and Widerface.
Our experimental results show that the proposed method achieves comparative accuracy comparing with state-of-the-art CNN-based face detection methods while reducing the model size by 
$3\times$ and memory-BW by $3-4\times$ 
comparing with 
one of the best real-time CNN-based object detector \emph{YOLO} \cite{YOLO_Redmon_2016_CVPR}. 
Our $8$-bit fixed-point TF-model provides additional $4\times$ memory reduction while keeping the accuracy nearly as good as the floating point model and achieves $20\times$ performance gain compared to the floating point model.
Thus the proposed model is amenable for embedded implementations and is generic to be extended to any number of categories of objects.   


\end{abstract}

%% file: introduction.tex
\section{Introduction} \label{sec:introduction}

Deep Convolutional Neural Network (CNN) based models are the current state-of-the-art for the task of object detection. 
The best methods for object detection aim to increase the accuracy on standard datasets.
They run on powerful GPUs that dissipate a huge amount of power. 
On the other hand, embedded processors and DSPs are great low-power solutions where the instruction sets benefit from fixed-point operations. 
For practical deployment of object detector on mobile devices, we need low-complexity CNN models that can run on embedded processors.
The algorithms need to leverage the fixed point operations without compromising the accuracy. 

In this paper, we propose \emph{LCDet}, a low-complexity object detector to address the above issues.  
We design and develop an end-to-end TensorFlow-based fully-convolutional deep neural network for object detection inspired by YOLO \cite{YOLO_Redmon_2016_CVPR}. The differences of the proposed network from YOLO are described in section \ref{Network Architecture}.

We choose face detection as a use-case due to its many practical applications in mobile phones, although the algorithm is generic enough for any number of classes. 
The detection pipeline of our TensorFlow-Slim based network requires a single forward pass through the network.
Evaluation results for the face detection performance on publicly available datasets such as FDDB \cite{fddbTech} and Widerface \cite{widerface_yang2016} show that the proposed method achieves comparative accuracy with respect to state-of-the-art CNN-based face detection methods while reducing the model size by 
$3\times$ and memory-BW by $3-4\times$ 
comparing with YOLO \cite{YOLO_Redmon_2016_CVPR}, one of the fastest DCN-based object detector. 
Additionally, we quantize the model by $8-$bit precision, which leads to additional $4\times$ memory reduction with almost no loss in detection accuracy. The $8-$bit quantization is one of the most important steps for a deployment in fixed-point architectures such as DSPs or dedicated convolution accelerators. 

We believe we report one of the first studies of $8-$bit quantization on TensorFlow models for object detection task that heavily uses \emph{regression}.
It is understood that $8$-bit quantization of floating point models that were trained for regression task are more prone to accuracy drop comparing with the models that were trained for classification task. Experimental results show that the highest detection accuracy with quantized model drops by less than $1-2\%$  comparing with the floating point model and achieves $20\times$ performance gain in terms of frame rate compared to the floating-point model.

The rest of the paper is organized as follows. We discuss the related work in section \ref{sec:prior_art}. We present the method, including the architecture, training, and model quantization in section \ref{sec:methods}. In section \ref{sec:results}, we report our results on face detection task and discuss relative complexity and accuracy. Finally, we conclude in section \ref{sec:conclusions}.

%% file: related_work.tex
\section{Related Work}
\label{sec:prior_art}

\subsection{CNN-based Object Detection}
Several papers propose ways of using deep convolutional networks for detecting objects  \cite{RCNN_girshick14CVPR,fast_RCNN_15,Faster_RCNN_RenHG015,SzegedyREA14, SzegedyREA14, Overfeat_SermanetEZMFL13, CRAFTCVPR16, Gidaris_2015_ICCV, ShallowNet_AshrafWIMK16}. 
Some approaches classify the proposal regions \cite{RCNN_girshick14CVPR,fast_RCNN_15} into object categories and some other recent methods \cite{Faster_RCNN_RenHG015,YOLO_Redmon_2016_CVPR,SSD_LiuAESR15}
unify the localization and classification stages. Detailed prior-art on object detectors and their speed-accuracy trade-off can be found in \cite{Speed_accu_obj_det_HuangRSZKFFWSG016}.

Precisely, the single stage detection pipeline of YOLO \cite{YOLO_Redmon_2016_CVPR} is extremely fast. YOLO is the first reported real-time CNN-based object detector model that runs with high-end GPUs. Its performance accuracy on PASCAL VOC \cite{PASCAL_VOC} dataset is comparable with state-of-the-art methods. 

Unlike YOLO, our model is fully-convolutional. Thus it is highly memory-efficient, computationally more effective and not restricted by input image resolution. 

\subsection{CNN Object Detection for Embedded Systems}
\noindent
The most accurate and best performing CNN-based models require high-end GPUs. There is a growing interest for developing specific hardware design \cite{modivius, Eyeriss_Chen_2016} including FPGAs, DSPs, custom vision chips and embedded GPUs for energy efficient CNN-based object-detection. 
Detailed algorithmic advancements and case-studies for CNN-algorithms for embedded systems can be found in \cite{Qiu_embedded_16, Embedded_CNN_obj_det_16}. 

The best-performing CNN-based object detection methods which run on real-time (on high-end GPUs), falls significantly short on embedded GPUs. For example, a supplier in surveillance camera market found out that even after replacing the back-end of YOLO from GoogleNet to a simpler CNN such as AlexNet, it's embedded implementation runs at most 5 frames per second on embedded GPUs. 
This motivates us to investigate on fully-convolutional low complexity object detector that can run real-time on embedded platforms.

TensorFlow is an open source framework and used by many developers to create their own AI-applications. Recently, \cite{Snapdragonblog} announced that  Snapdragon 835 \cite{Snapdragon835} includes TensorFlow-optimized Hexagon 682 DSP. This DSP architecture and others in this family are designed to process certain features more quickly and at lower power than a CPU or GPU. 
Our proposed TensorFlow-based model exploits the advantages of similar architecture and is useful for real-time object detection tasks on these platforms. 

\subsection{CNN-based Face Detection}

We choose face detection as an application and evaluate \emph{LCDet}, the proposed object detector, for this task.  
As per the FDDB evaluation server \cite{fddbTech}, the state-of-the-art face detection methods  are based on convolutional neural networks ~\cite{faceness, Conv3D, UnitBox, Xiaomi, FD-RCNN, DeepIR, Qin_2016_CVPR}. Yang \etal presented a neural network which combines feature responses regarding facial parts~\cite{faceness}. Li \etal presented an integrated method of neural networks and 3D face model~\cite{Conv3D}. Yu \etal modified VGG-16 networks, and also proposed intersection over union (IoU) loss layer~\cite{UnitBox}. Recently presented works are based on faster R-CNN~\cite{Faster_RCNN_RenHG015, Xiaomi, FD-RCNN, DeepIR}. 
Our model achieves comparable quality with Faster-RCNN based methods. Additionally, our method meets all the requirements for real-time embedded applications while the above other CNN-based face detection methods can not achieve real-time performance in embedded platforms. 


%% file: methods.tex
\section{Methods}
\label{sec:methods}

\subsection{Network Architecture} \label{Network Architecture}

\begin{figure*}[!t]
\centering
\fbox{\includegraphics[width=2.96in]{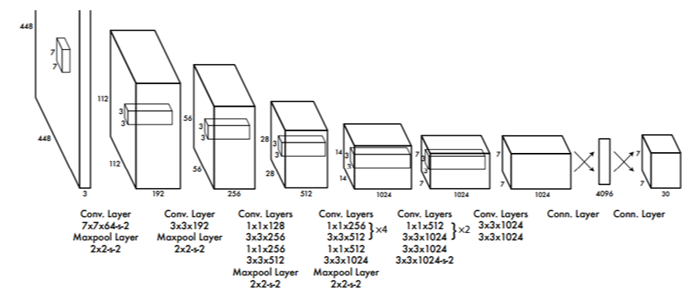}}
\fbox{\includegraphics[width=3in]{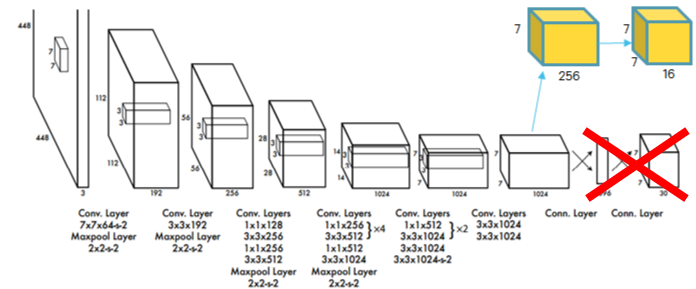}}
\caption{YOLO vs proposed network. Output channel's value $16$ corresponds to $1$ class-conditional and $1+4$ confidence and coordinates for each of the $B=3$ boxes. All Leaky ReLU activations are replaced by ReLU in the final model of LCDet.}
\label{fig:architecture}
\end{figure*}

Our proposed model is inspired by YOLO \cite{YOLO_Redmon_2016_CVPR} which adopts a single-pass detection pipeline combining bounding box localization and classification by a single network.
Layers connectivity differences between YOLO and LCDet is outlined in Figure \ref{fig:architecture}. The last two fully-connected layers of YOLO are replaced by fully-convolutional layers in the proposed model.

Other differences are described below. 
Unlike the LeakyReLU non-linearity in YOLO, we apply ReLU activations in all but last layer. 
Additionally, for the final layer of output, we apply different activations on classification (softmax) output, confidence (sigmoid) score, and localization (no activation) outputs. 
YOLO doesn't apply any non-linearity in the final layer.
From the layers connectivity perspective, the back-end of the CNN architecture is almost similar to YOLO; however, \emph{LCDet} can work on any input image resolution by virtue of being fully-convolutional. 

Let's suppose, the convolutional layer right before the first fully-connected layer in YOLO is called the final feature map of spatial size $W_f\times H_f$. Here, $W_f$ and $H_f$ denote the number of grid centers along the horizontal and vertical axes. 

From the same feature layer, the proposed model connects to the final convolutional layer that outputs in a spatial grid-like pattern ($W_f\times H_f\times Channels$) as shown in Figure \ref{fig:architecture}.
Each grid center is associated with $C$ class probabilities, $1$ confidence score, and $4$ scalar values of coordinates for each of the B ($=3$) possible bounding boxes. Similar to YOLO, the confidence score is the predictor for Intersection-over-Union with the ground truth bounding box. 
Finally, we employ Non-Maximum suppression (NMS) for keeping top bounding boxes. During the inference, the detection pipeline consists of a single forward pass through the network.

\subsection{Training Methodology} \label{training_details}

Unlike Faster R-CNN \cite{Faster_RCNN_RenHG015}, which deploys a 4-step alternating
training strategy to train Region Proposal Network (RPN) and detector network,
our detection network can be trained
end-to-end, similarly to YOLO \cite{YOLO_Redmon_2016_CVPR}.
We apply a multi-part object detection loss as described in (equation \ref{eqn:obj_det_loss}) similar to YOLO.

\small
\begin{equation} \label{eqn:obj_det_loss}
\begin{aligned}
\mbox{loss}
= \lambda_{coord}\sum^{S^2}_{i=0}\sum^{K}_{j=0}\mathbbm{1}^{obj}_{ij}\big(\mathit{x}_i - \hat{\mathit{x}}_i\big)^2 + \big(\mathit{y}_i - \hat{\mathit{y}}_i\big)^2 + \\
\lambda_{coord}\sum^{S^2}_{i=0}\sum^{K}_{j=0}\mathbbm{1}^{obj}_{ij}\big(\sqrt{w_i} - \sqrt{\hat{w}_i}\big)^2 + \big (\sqrt{h_i} - \sqrt{\hat{h}_i} \big)^2 +\\
\sum^{S^2}_{i=0}\sum^{K}_{j=0}\mathbbm{1}^{obj}_{ij}(\mathit{C}_i - \hat{\mathit{C}_i})^2 +\\
\lambda_{noobj}\sum^{S^2}_{i=0}\sum^{K}_{j=0}\mathbbm{1}^{noobj}_{ij}\big(\mathit{C}_i - \hat{\mathit{C}}_i\big)^2 +\\
\sum^{S^2}_{i=0}\mathbbm{1}^{obj}_{i}\sum_{c \in classes}\big(p_i(c) - \hat{p_i}(c)\big)^2
\end{aligned}
\end{equation}
\normalsize

where $\mathbbm{1}^{obj}_{i}$ denotes 
if the object appears in cell $i$ 
and $\mathbbm{1}^{obj}_{ij}$ 
denotes that $j$th bounding box predictor 
in cell $i$ 
is \emph{responsible} for that prediction. 
The loss function penalizes classification 
and localization error differently 
based on presence or absence of an object 
in that grid cell. 
$x_i, y_i, w_i, h_i$ corresponds 
to the ground truth bounding box center coordinates, width and height for objects in grid cell (if it exists) and $\hat{x_i}, \hat{y_i}, \hat{w_i}, \hat{h_i}$ stand for the corresponding predictions. 
$C_i$ and $\hat{C_i}$ denote confidence score of \emph{objectness} at grid cell $i$ for ground truth and prediction.  
$p_i(c)$ and $\hat{p_i}(c)$ 
stand for conditional probability 
for object class $c$ at cell index $i$ 
for ground truth and prediction respectively.  
We use similar settings for YOLO's object detection loss minimization and use values of $\lambda_{coord} = 5$ and $\lambda_{noobj} = 1.$.\\

We additionally apply sigmoid activation on the prediction of confidence score.
Confidence score should be in [0,1] as it is ideally the IOU predictor. 
We employ the softmax on class prediction. However, for the special case of single-class detection in YOLO-style, we employ sigmoid activation on $1$ class prediction output from the network. 

The proposed model uses $448 \times 448$ frames as input 
while training 
and regresses on category types and locations of possible objects
at each one of $S \times S$ non-overlapping grid cells.
(The model is capable of using any resolution image as an input)
For each grid cell, 
the model outputs class conditional probabilities as well as $K$ bounding boxes
and their associated confidence scores. 
As in YOLO, we consider a \emph{responsible} bounding box for a grid cell to be the one among the $K$ boxes for which the predicted area and the ground truth area shares the maximum Intersection Over Union. 
During training, we simultaneously optimize classification and localization error 
(equation \ref{eqn:obj_det_loss}).
For each grid cell, 
we minimize the localization error 
for the \emph{responsible} bounding box 
with respect to the ground truth 
only when an object appears in that cell.  


\subsection{Detection-Specific Layers} \label{det_layer}

From the feature layer of size $W_f\times H_f \times Ch_f$,  YOLO \cite{YOLO_Redmon_2016_CVPR} employs two fully-connected layers. For simplicity, we denote these two layers together as YLDet. We denote the last two convolutional layers of the proposed model by ConvDet. 

\begin{table}[hbt!]
\centering
\small
\begin{tabular}{llll} 
& \multicolumn{1}{c}{\textbf{RP}}
& \multicolumn{1}{c}{\textbf{cls}}
& \multicolumn{1}{c}{\textbf{\# Parameters}}
\\
\toprule
\\
\textbf{RPN} & \cmark &  \xmark & $Ch_fK(5+C)$
\\
\textbf{YLDet}  & \cmark & \cmark & $F_{fc1}(W_fH_fCh_f+W_oH_o(C+5K))$
\\
\textbf{ConvDet}   & \cmark &\cmark & $F_wF_hCh_{d1}(Ch_f + (C+5K))$
\\
\bottomrule
\\
\end{tabular} 
\caption{
Comparison between RPN, ConvDet and YLDet. RP stands for Region Proposals, \emph{cls} denotes classification. 
}
\label{tab:detection-specific-net}
\end{table}

YOLO works with input feature map size of $7\times 7 \times 1024$. $F_{fc1}=4096$, $C=20$, $W_0=H_0=7$. Thus the number of parameters in \emph{YLDet} is about $269\times 10^6$. The first convolutional layer in the 
ConvDet has $Ch_{d1}=256$ parameters. For same feature map size and number of output grid centers, ConvDet only requires $2.3\times 10^6$ parameters, which is $115\times$ less than YOLO.

\subsection{Quantized Model}

Often times, DSPs or dedicated convolution accelerators operate on fixed point instruction set. There exists literature on fixed point models for embedded systems \cite{Quan_cls_mobile_WuLWHC15, Ristretto_Gysel16} for classification task. 
It is well-known that $8-$bit models 
\cite{TF_quant2} perform as good as the floating point model for classification \cite{FixedPoint_forCNN}.
The justification for the high accuracy in low-precision modes comes from the fact that the final activation is a probability \ie in [0,1] intervals, can be represented with an unsigned number without any concern on scaling. However, we are not aware of many published reports on the study of quantization for object detection task that regresses coordinates of all objects. 

For each layer of LCDet, we convert the 32-bit floating point parameters to 8-bit fixed point parameters via \cite{TF_quant1, quantization}. The entire object detection model can work in 8-bit fixed point implementation without going back and forth from floating and fixed point after the accumulation in each convolutional layer.  
Although literature exists for training with low precision mode \cite{trainLow_prec_CourbariauxBD14}, we perform quantization only for the inference. 
Since 
training is performed off-line, it is 
reasonable and more practical
to quantize the trained model for inference.

To quantize the 32-bit floating point to the 8-bit fixed point, we first store minimum and maximum value at a layer. Then, we quantize the relative value to the linearly distributed closest integer in [0,255].
\begin{equation}
w_q = \Big[255 \frac{w_f - w_{min}}{w_{max} - w_{min}} \Big]
\end{equation}
where $w_q$, $w_f$, $w_{min}$, and $w_{max}$ represents quantized variable, variable in floating point, minimum, and maximum. $[\cdot]$ represents rounding to the closest integer.  

Although we know of no fundamental mathematical reason as to why the low precision mode works well without low-precision training, 
we see from our experimental results that regression models (such as our LCDet) also works well for low-precision inference. 
Quantized LCDet is as good as the floating-point model in terms of detection accuracy for the face detection application in general.

%% file: results.tex
\section{Experimental Results} \label{sec:results}

In this section, we present the performance of the proposed algorithm with floating point model as well as the associated $8$-bit quantization on commonly used face databases such as FDDB benchmark \cite{fddbTech} and Widerface \cite{widerface_yang2016} validation split. 
We also provide
an analysis of the proposed method's performance in terms of speed, complexity and memory requirements.
Although, face detection is studied as a use case, the network does not use any face-specific processing such as facial parts or attributes. LCDet is a general purpose object detector.   


We use transfer learning by first converting the weights for detecting 20-category PASCAL objects 
\cite{PASCAL_VOC} from DarkNet \cite{Darknet} library to a TensorFlow checkpoint, called \emph{YOLO PASCAL TF}. 
For face detection with baseline YOLO \cite{YOLO_Redmon_2016_CVPR}, we first restore parameters from the \emph{YOLO PASCAL TF} checkpoint for all except the last layer. 
We then fine-tune it for the 1-class (only face) object detection task. 
For LCDet, we restore parameters for all but last two layers from the \emph{YOLO PASCAL TF}. We then initialize the last two layers with random weights and train the proposed LCDet for single-class object detection. 
Also, we replace all LeakyReLU activations from YOLO architecture to ReLU activation after each convolution except the final output layer. Readers are referred to section \ref{training_details} for the activations in the last layer. 

We adopt data augmentation techniques such as scaling and object-centric cropping to minimize overfitting. 
In our experiments, we use initial learning rate of $10^{-5}$, the minibatch size of $32$, and $8K$ epochs. 
We use Adam \cite{Adam_KingmaB14} 
optimizer for training. We used NVIDIA 
Tesla K40
GPUs for training and testing experiments. We also trained similar models with batch normalization for the convolutional layers. The training appeared to converge early, but that didn't yield any improvement in the detection accuracy. In order to minimize the number of parameters, we go with models without batch normalization. 

\subsection{Dataset}

FDDB \cite{fddbTech} is a benchmark dataset for face detection in unconstrained settings. It contains $2,845$ images with a total of $5,171$ faces. The dataset provides fixed partitioning of $10$ folds. 
WIDER FACE is a larger dataset \cite{widerface_yang2016} for face detection. It consists of $32,203$ images with $393,703$ faces. The dataset is organized in $61$ event classes. For each event class, $40\%$, $10\%$, and $50\%$ of data were selected for training, validation, and testing. The ground truth for test split is not disclosed. 
In our experiments, we scaled each image to a pre-determined size. For gray-scale images, we duplicate the single channel three times to make them images with three channels.

\subsection{Detection Accuracy}

\begin{figure}[!t]
\centering
\includegraphics[width=1.0\linewidth]{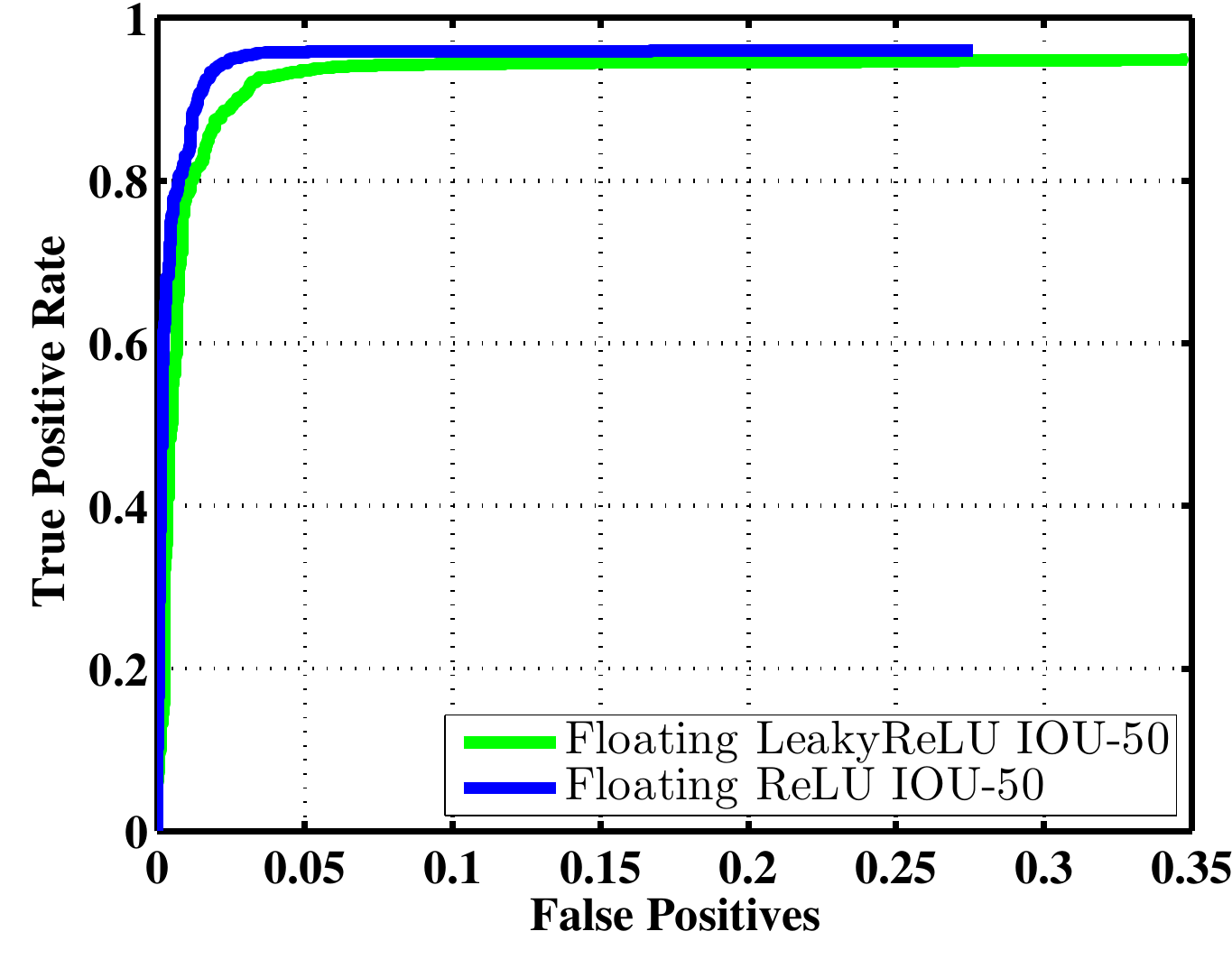}
\caption{Performance of LCDet on FDDB for LeakyReLU vs ReLU model with discrete core metric.}
\label{fig:LeakyReLUvsReLU}
\end{figure}

We first evaluate LCDet for two different nonlinearities such as Leaky ReLU and ReLU at each convolutional layers.  
Although the model has been initialized from YOLO PASCAL TF which was trained with Leaky ReLU activations. After finetuning for face detection task with two different models with two different activations, we find ReLU activation performs better than leaky ReLU.
The performance of LCDet with LeakyReLU vs ReLU has been shown in Figure \ref{fig:LeakyReLUvsReLU}. 
Next we evaluate the performances of LCDet and its 8-bit quantized model on FDDB dataset. The models are trained on FDDB images. This allows us to investigate the performance gap of floating and fixed point models independent of the other components in the whole system. 



\begin{figure}[!t]
\centering
\includegraphics[width=1.0\linewidth]{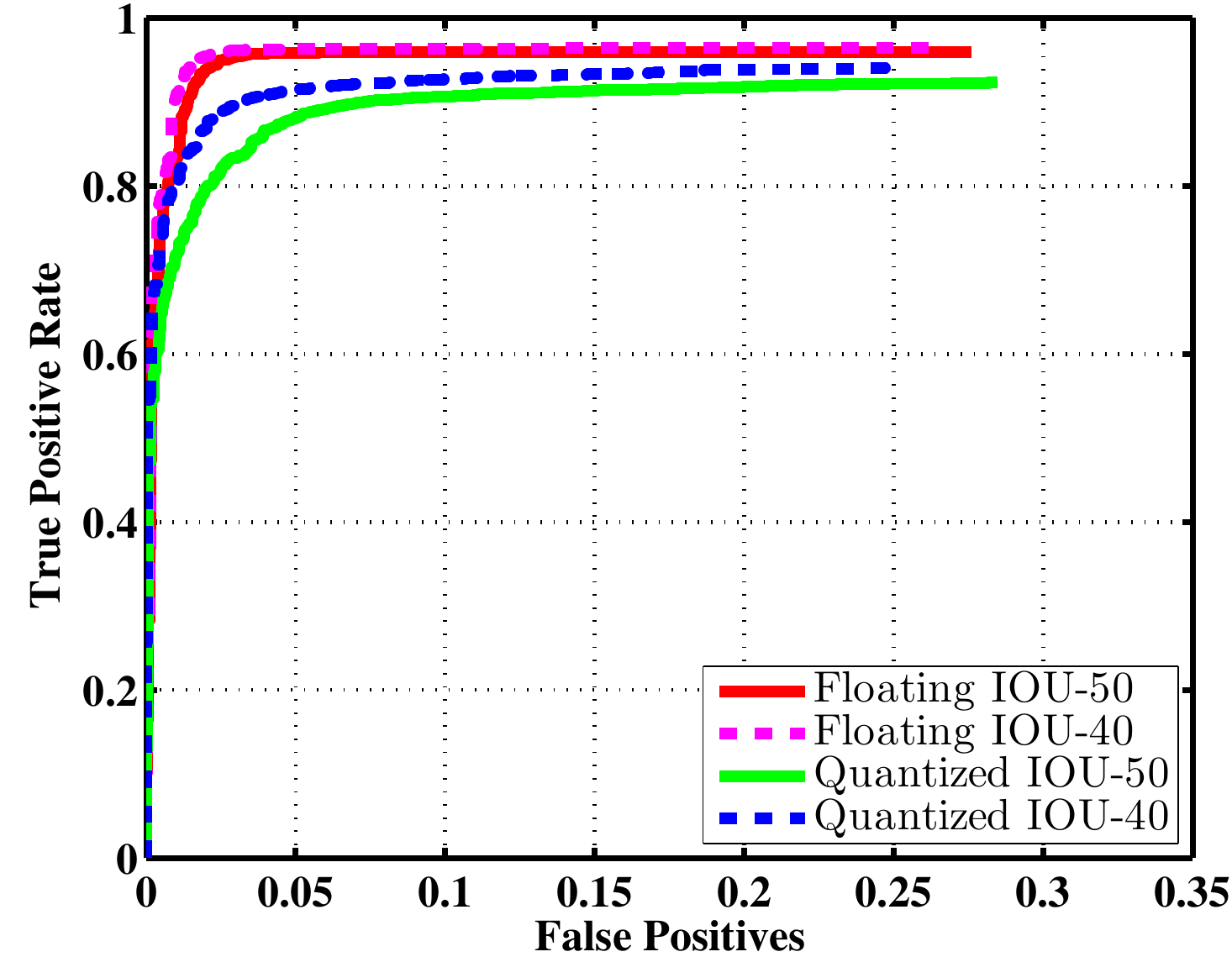}
\caption{Performance of Fixed vs Floating point models (ReLU) on FDDB with discrete score metric. Effects of quantization on regression is better understood with relaxed detection criteria in terms IOU going down from 50\% to 40\%. Although the floating point model achieves a little improvement, the fixed point model achieves 5\% improvement in true positives. As expected, regression of box coordinates appear to be affected by quantization significantly. }
\label{fig:ReLU_Fixed_Floating}
\end{figure}

\begin{figure}[!t]
\centering
\includegraphics[width=1.0\linewidth]{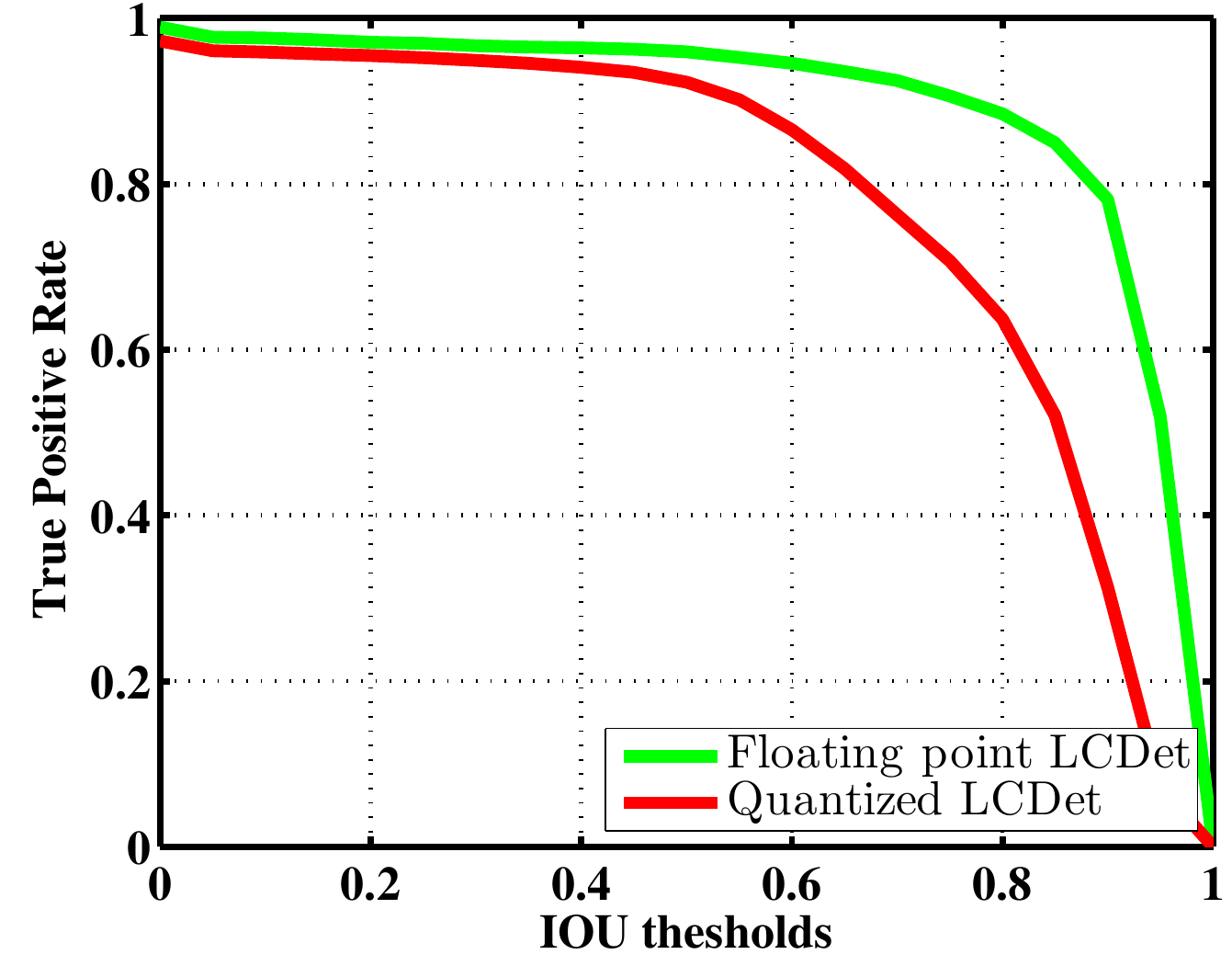}
\caption{Performance comparison between floating and fixed point models (ReLU) at different IoU thresholds on FDDB.}
\label{fig:TPvsIoU}
\end{figure}

Figure 
\ref{fig:ReLU_Fixed_Floating}
shows the performance of the floating point model and the 8-bit quantized model by the TP-FP curve with discrete score evaluation method per \cite{fddbTech}. Solid curve denotes the performance at standard detection crietria \ie $50\%$ Intersection over Union (IoU) with ground truth. Dotted lines denote less strict but practical detection criteria \ie $40\%$ IoU with ground truth boxes. 
Regressed coordinate locations from the fixed point model suffer from higher deviation from the floating point model prediction. This effect becomes more evident when we increase the IoU criteria for detection. As we see in Figure \ref{fig:TPvsIoU}, for more relaxed IoU criteria such as $40\%$ IoU, floating vs fixed point model exhibit similar detection performances. However, for stricter detection criteria such as $60\%$ IoU or higher, the performance of the fixed-point model appears to drop significantly, especially for the lower false positive regions on the TP-FP curves.

Mobile devices use face detection for several face based quality enhancement processing such as auto-exposure or auto-focus. Any false detection should be highly penalized as their consequences are more expensive. On the other hand, if the detected box overlaps with the actual face by little less than $50\%$, certain end use cases that use face detection bounding box as input can still function with similar performance. In less strict IoU operating region, LCDet fixed point model is regarded as good as the the LCDet floating point model. Figure \ref{fig:TPvsIoU} shows that the detection rate in those operating points (upto 45\% IoU) is similar for floating and fixed LCDet models.

\begin{figure}[!t]
\centering
\includegraphics[width=0.85\linewidth]{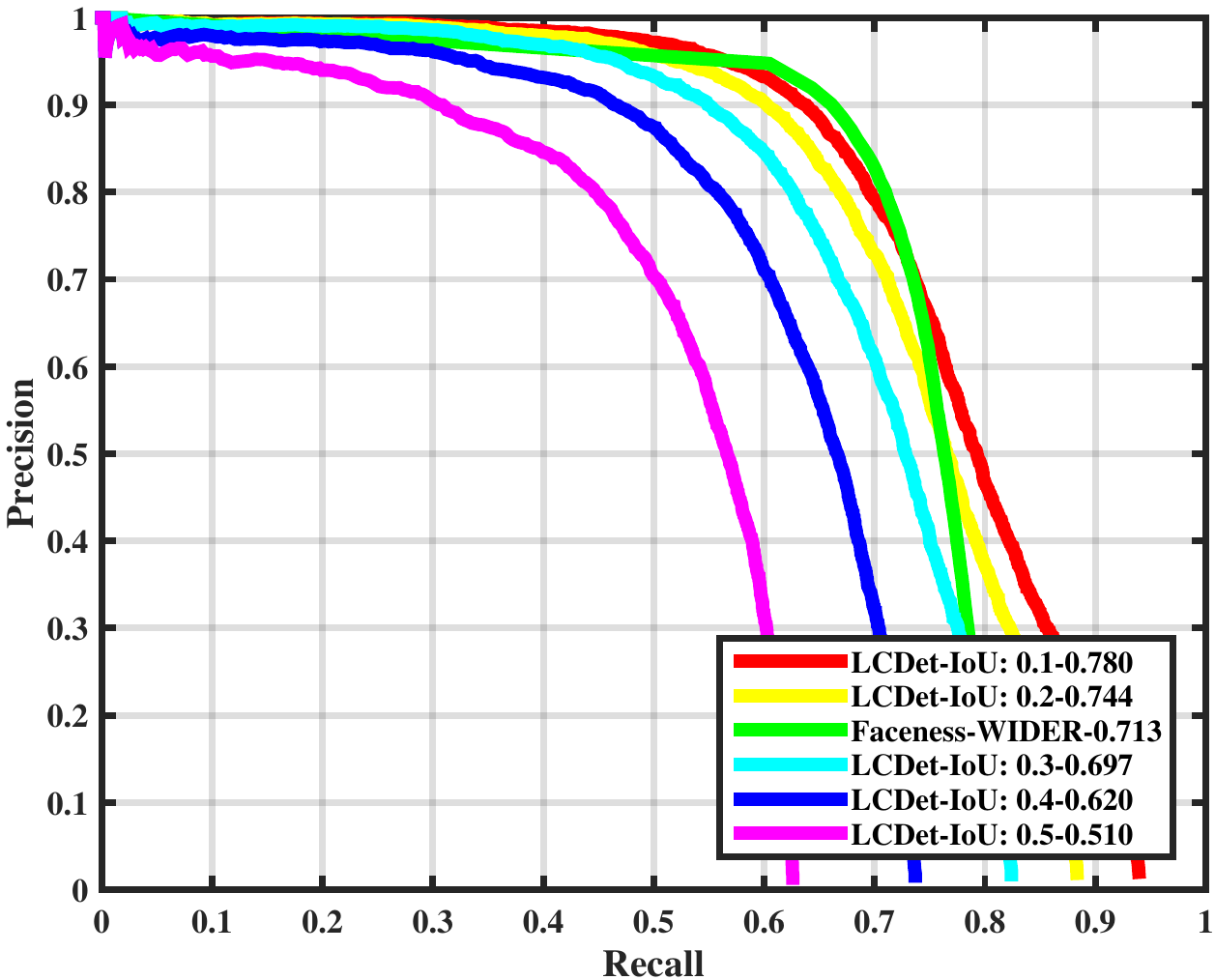}
\caption{Precision-Recall performance on Widerface Validation set with relaxed IoU criteria.}
\label{fig:Wider_val_IOU}
\end{figure}

Next, we train LCDet on the Widerface training split. The Widerface has about $20\times$ more faces than FDDB. some of the faces are extremely small. We evaluate the performance of LCDet on the Widerface validation split using the provided evaluation toolbox.
One of the current limitation of the YOLO-style training is that it assumes at most one ground truth object at each grid location, although it can predict up to $k$ objects per grid. For a training image of size $448\times 448$ means $7\times 7$ grids, thus can exploit only $49$ ground truth objects. On the other hand, the Widerface has more than $100$ ground truth faces in at least $200$ training images. YOLO-style training could not use all the ground truth available. In such cases, we used the ground truth with highest area per grid location. On the contrary, faster-RCNN type network can use all possible ground truth objects. As shown in Figure \ref{fig:Wider_val_IOU}, the model needs improvement for the localization accuracies especially for small objects those are present in Widerface. As the IoU criteria is getting relaxed, the model approaches comparable or even better accuracy than other state-of-the-methods such as Faceness \cite{faceness}.   


\subsection{Complexity and Memory-BandWidth analysis}

We first convert the darknet \cite{Darknet} YOLO implementation to TensorFLow-Slim based implementation. 
We leveraged the weights from the darknet \cite{Darknet} for all of the $24$ convolutional layers and first fully connected layer. 
The number of output nodes in the final detection layer is  $W_f\times H_f \times (C + K\times 5)$. For face detection task, $C=1$, and we use $K=3$ for all our experiments. 
Then using the mentioned training methodology, we fine-tune all layers for the face detection task. 
Table \ref{tab:Comparative-performances} demonstrates the performance and accuracy of these models along with some of the other recent models on powerful GPUs. 
\begin{table*}[hbt!]
\centering
\normalsize
\begin{tabular}{lllll} 
& & &  
\\
& \multicolumn{1}{c}{\textbf{Model}}
&
& \multicolumn{1}{c}{\textbf{Inference}}
&
\\
& \multicolumn{1}{c}{\textbf{Size}}
& \multicolumn{1}{c}{\textbf{FLOPs}}
& \multicolumn{1}{c}{\textbf{Speed}}
& \multicolumn{1}{c}{\textbf{Max}}
\\
& \multicolumn{1}{c}{\textbf{(MB)}}
& \multicolumn{1}{c}{\textbf{$\times 10^9$}}
& \multicolumn{1}{c}{\textbf{(FPS)}}
& \multicolumn{1}{c}{\textbf{TP}}
\\
\toprule
\textbf{LCDet} & 250 &  20 & 17.55 &  93.0 \\
YOLO* & 1126  & 20.1 & 15.44 & 85.0
\\
\midrule
Faster-RCNN+VGG16 & 485 
& 
98
& 
4.61
\cite{fasterRCNN_TF} 
& 96.1 \cite{FD-RCNN} \\
SSD300+VGG16\cite{SSD_LiuAESR15,SSD_TF}  & 105 & 31.6  & 17.97\cite{SSD_TF} & NA
\\
\\
\bottomrule
\\
\end{tabular} 
\caption{
LCDet vs other methods. Inference speed on NVidia Tesla K40. YOLO* is our TF-implementation for Face Detection. Max TP is the highest true positive rate value achieved in FDDB. SSD and Faster R-CNN running time use TF-SSD implementation \cite{SSD_TF} and TF Faster RCNN implementation \cite{fasterRCNN_TF} respectively in different implementation platforms.
}
\label{tab:Comparative-performances}
\end{table*}


\begin{table*}[hbt!]
\centering
\begin{tabular}{llllll} 
& & & \multicolumn{1}{c}{\textbf{Activation}}
\\
& \multicolumn{1}{c}{\textbf{Model}}
& 
& \multicolumn{1}{c}{\textbf{Memory}} 
\\
& \multicolumn{1}{c}{\textbf{Size}}
& \multicolumn{1}{c}{\textbf{OPs}}
& \multicolumn{1}{c}{\textbf{Footprint}}
\\
& \multicolumn{1}{c}{\textbf{(MB)}}
& \multicolumn{1}{c}{\textbf{$\times 10^9$}}
& \multicolumn{1}{c}{\textbf{(MB)}}
\\
\toprule
Quantized YOLO & 281 &  20.1 
& 333 
\\
\textbf{Quantized LCDet} & 62 &  20.0 
& 88 
\\
\bottomrule
\\
\end{tabular} 
\caption{
Performance analysis of Fixed-point LCDet in terms of 
OPs, Memory Activation Footprints 
}
\label{tab:Comparative-performances-on-DSP}
\end{table*}


The detection-specific module of LCDet uses two convolutional layers. The first one has $4096$ kernels of size $3\times3$ and the second one has $16$ kernels of $1\times1$ size each. 
Irrespective the backend feature-extractor network, 
LCDet has $115\times$ fewer parameters for only the detection part comparing with YOLO as described in sec \ref{det_layer}. 

\begin{figure}[!t]
\centering
\includegraphics[width=1.0\linewidth]{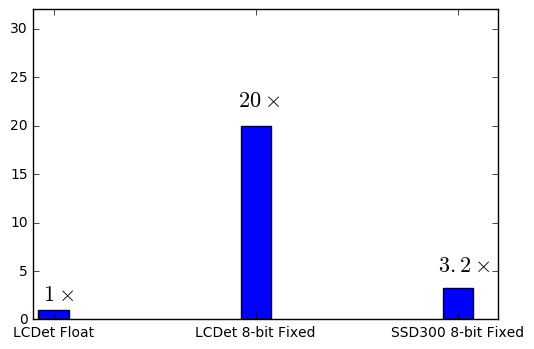} 
\caption{Frame Rate Improvement for Fixed Point Model}
\label{fig:framerate}
\end{figure}
\begin{figure}[!t]
\centering
\includegraphics[width=1.0\linewidth]{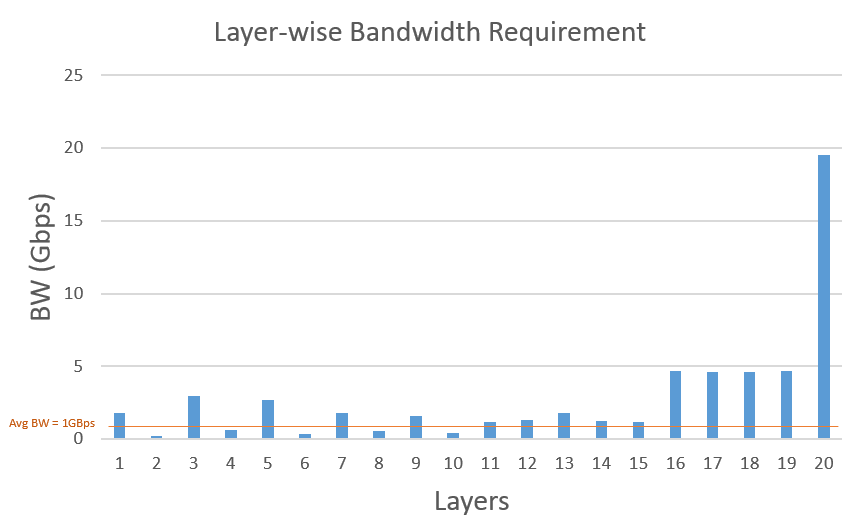}
\caption{Layer-wise Bandwidth Requirement for LCDet-8bit-fixed Point Implementation}
\label{fig:instantenousbw}
\end{figure}
\begin{figure}[!t]
\centering
\includegraphics[width=1.0\linewidth]{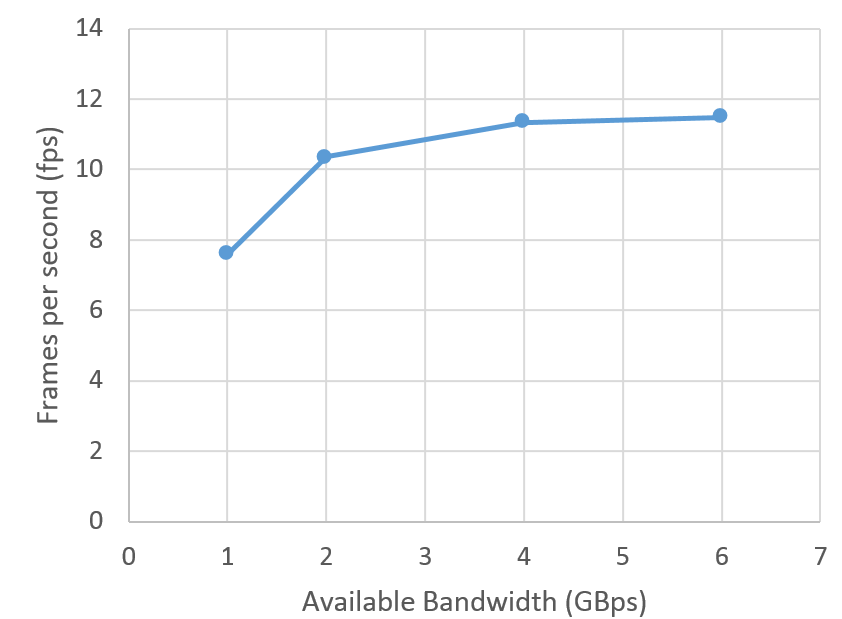}
\caption{Effect of available DDR BW on Frame Rate}
\label{fig:fpsvsbw}
\end{figure}

Next, we analyze and compare the performance of proposed method with respect to the state-of-the-art deep neural network-based object detector simulated on a commercially available Snapdragon platform such as in \cite{Snapdragon835}. Hexagon DSP includes fixed-point vector extensions which make it an attractive computing unit for computer vision applications and provides performance per power compared to CPU and GPUs on mobile platforms. We quantized our baseline YOLO model and also quantized the proposed LCDet model, and compare their relative model sizes and activation memory footprints for the same input resolution size as shown in Table \ref{tab:Comparative-performances-on-DSP}. 
In Figure \ref{fig:framerate}, we compare the achievable frame rate of the following methods: LCDet-float, LCDet-8bit-fixed, and SSD300-8bit-fixed. In a fixed-point implementation, activations and weights are implemented in 8-bits and mapped to vector extension of DSP, whereas in float implementation vector extension is not utilized. By bringing down the model size and bandwidth (BW) per layer, we achieve close to the $20\times$ increase in frame rate with respect to floating point implementation. The average DDR bandwidth that our quantized model requires is roughly 1 Gbps, whereas the instantaneous BW has a wider range reaching close to 20 Gbps for some layers as shown in Figure \ref{fig:instantenousbw}. Typically, the DDR BW is throttled when multiple applications are run on the embedded systems and frame rate degrades because of stringent BW constraints. This is depicted in Figure \ref{fig:fpsvsbw}, where the frame rate drops as DDR BW decreases from $6$ Gbps to $1$ Gbps.

\subsection{Visual Results}

\begin{figure}[!t]
\centering
\includegraphics[width=1.0\linewidth]{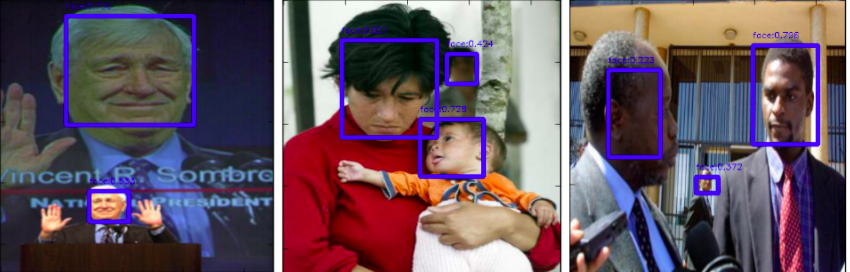}
\caption{Detected faces of different scales on FDDB.}
\label{fig:different_scales}
\end{figure}

\begin{figure}[!t]
\centering
\includegraphics[width=1.0\linewidth]{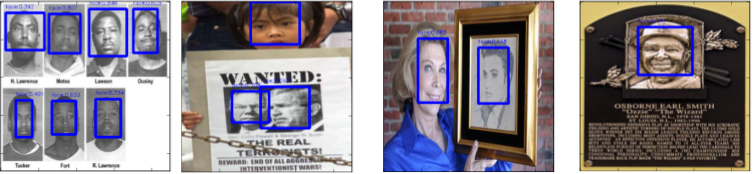}
\caption{Faces detected in black and white images on FDDB.}
\label{fig:BW}
\end{figure}

\begin{figure}[!t]
\centering
\includegraphics[width=1.0\linewidth]{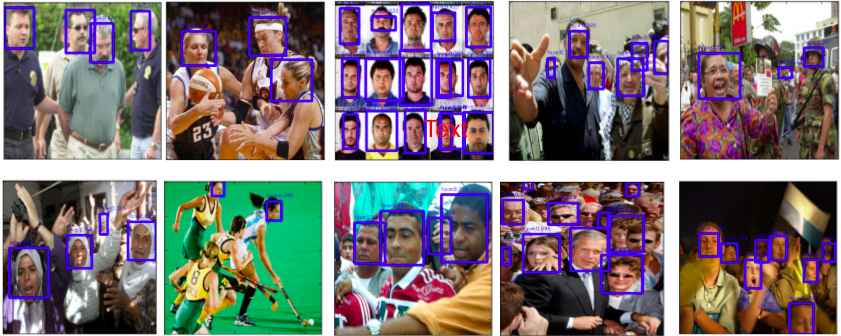}
\caption{Multiple faces of frontal and side profiles on FDDB.}
\label{fig:multiple_faces}
\end{figure}

\begin{figure}[!t]
\centering
\includegraphics[width=1.0\linewidth]{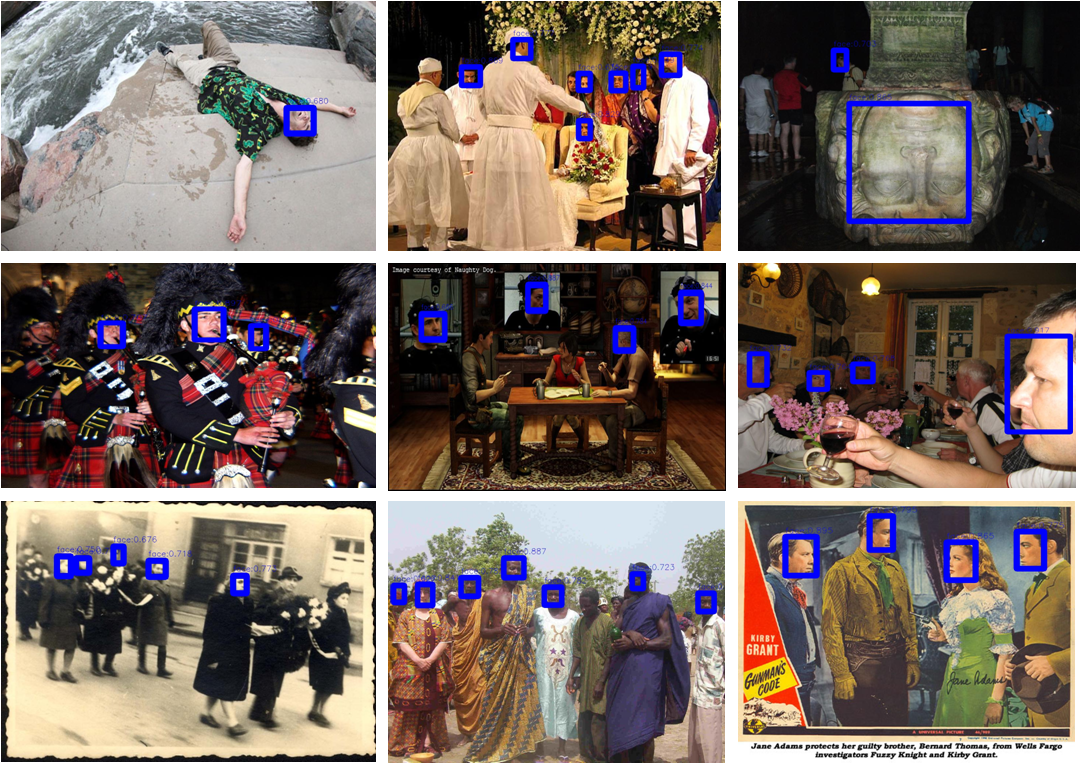}
\caption{Successful face detection results of LCDet on challenging Widerface Validation images containing difficult examples in pose variation, illumination changes, photograph styles, and different sizes.}
\label{fig:wider_successful}
\end{figure}

We show visual detection results of the face detection performed by the proposed LCDet. Detected faces are marked as \emph{blue} rectangle. Figure \ref{fig:different_scales} to Figure \ref{fig:multiple_faces} demonstrate face detection results on FDDB dataset.
These figures show that LCDet detects multiple faces of different sizes and poses accurately for a variety of illumination and scale changes.
Some of the difficult examples from the Widerface validation dataset are shown in Figure \ref{fig:wider_successful}. In general, LCDet performs well in detecting faces.  

\begin{figure}[!t]
\centering
\includegraphics[width=1.0\linewidth]{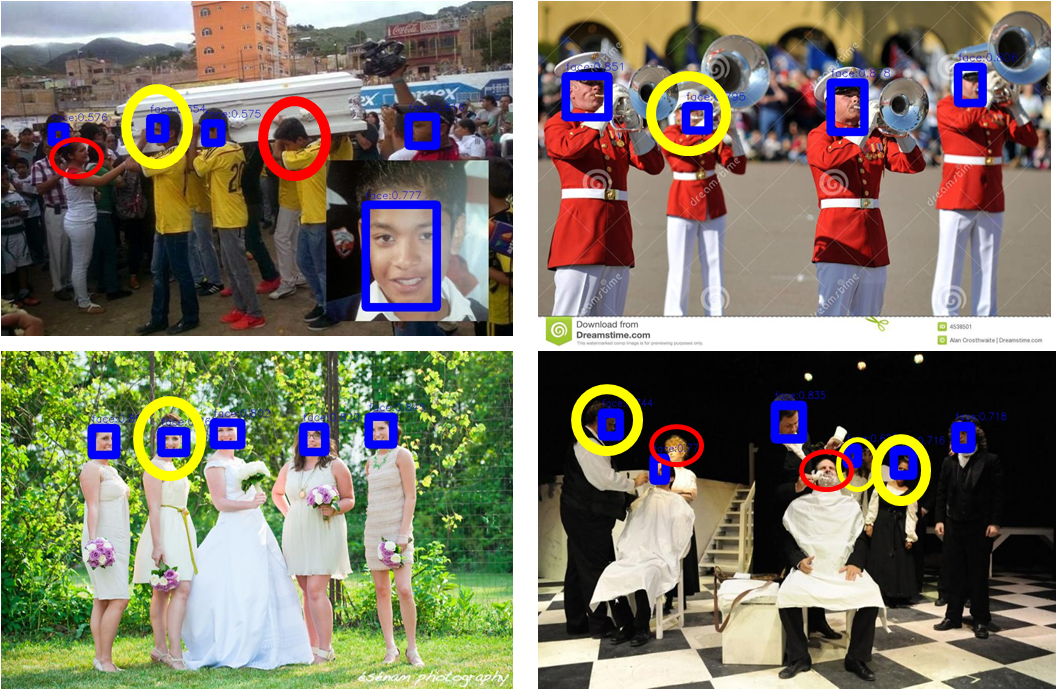}
\caption{Localization challenges of LCDet on Widerface validation images. Faces marked as the \emph{yellow} regions are considered false positives as per 50\% IoU criteria, but true positive for more relaxed IoU criteria. The regions marked in \emph{red} show missed detections.}
\label{fig:wider_challenges}
\end{figure}

Current limitation of the model is that it struggles in tightly localizing  small objects in close proximity. Figure \ref{fig:wider_challenges} demonstrates some of the examples where localization might have failed as per strict $50\%$ IoU criteria (marked as \emph{yellow} regions), however the detected faces are not false positives for further face-based processing pipeline. The regions marked in \emph{red} shows missed detections.

%% file: conclusions.tex
\section{Conclusions}
\label{sec:conclusions}

We propose LCDet, a low-complexity fully-convolutional neural network for object detection amenable for embedded deployment. This is a unified localization and classification model inspired by \cite{YOLO_Redmon_2016_CVPR} that bypasses the object proposals bottleneck. LCDet performs comparably with state-of-the-art CNN-based face detection methods on FDDB, while being one of the most computationally effective method. 
We additionally perform $8$-bit quantization on this TF-slim based LCDet model, and report one of the first analysis of quantized model for regression. The quantized LCDet model performs as good as floating point model and reduces the memory footprint by $4\times$.
Quantization makes this model apt for implementation in DSPs, or dedicated convolution accelerators.
Although, face detection is studied as a use case here, the network is not optimized for face-specific detection only. It is easily expendable for detecting any other categories of objects.  

We are aware of the very recent work YOLO9000 \cite{YOLO_Redmon_2017_CVPR} that has become the state-of-the-art on standard object detection datasets. YOLO9000 that is an improved version of YOLO. Empirically, the accuracy of LCDet lies between YOLO \cite{YOLO_Redmon_2016_CVPR} and YOLO9000 \cite{YOLO_Redmon_2017_CVPR}.
On the other hand, another recent work SqueezeDet \cite{SqueezeDet_WuIJK16} on low-complexity CNN achieves state-of-the-art performance on KITTI object detector. SqueezeDet appears to be the smallest object detector by virtue of powerful but small backend network of SqueezeNet \cite{SqueezeNet_IandolaMAHDK16}.  
At the time of writing, there is no reported performance comparison for all these on the same dataset. 
As a future work, we will evaluate the relative  performance of our proposed model comparing with these methods. It is also interesting to explore SqueezeNet as the backend and study performance of quantization.



%% file: ack.tex
\section*{Acknowledgments}

The authors would like to thank Vikram Gupta and Rakesh Nattoji Rajaram for extensive assistance and insightful comments.